\pdfoutput=1
\documentclass[sigconf]{acmart}

\makeatletter
\def\@ACM@checkaffil{
    \if@ACM@instpresent\else
    \ClassWarningNoLine{\@classname}{No institution present for an affiliation}%
    \fi
    \if@ACM@citypresent\else
    \ClassWarningNoLine{\@classname}{No city present for an affiliation}%
    \fi
    \if@ACM@countrypresent\else
        \ClassWarningNoLine{\@classname}{No country present for an affiliation}%
    \fi
}
\makeatother

\usepackage{multirow}
\usepackage{balance} 

\AtBeginDocument{%
  }

\copyrightyear{2023}
\acmYear{2023}
\setcopyright{acmlicensed}\acmConference[MM '23]{Proceedings of the 31st ACM International Conference on Multimedia}{October 29-November 3, 2023}{Ottawa, ON, Canada}
\acmBooktitle{Proceedings of the 31st ACM International Conference on Multimedia (MM '23), October 29-November 3, 2023, Ottawa, ON, Canada}
\acmPrice{15.00}
\acmDOI{10.1145/3581783.3611769}
\acmISBN{979-8-4007-0108-5/23/10}

\begin{document}

\title{Symmetrical Linguistic Feature Distillation with CLIP for Scene Text Recognition}

\author{Zixiao Wang}
\email{wzx99@mail.ustc.edu.cn}
\affiliation{%
  \institution{University of Science and Technology of China}
}

\author{Hongtao Xie}
\email{htxie@ustc.edu.cn}
\affiliation{%
  \institution{University of Science and Technology of China}
}

\author{Yuxin Wang}
\email{wangyx58@mail.ustc.edu.cn}
\authornote{Corresponding Author.}
\affiliation{%
  \institution{University of Science and Technology of China}
}

\author{Jianjun Xu}
\email{xujj1998@mail.ustc.edu.cn}
\affiliation{%
  \institution{University of Science and Technology of China}
}

\author{Boqiang Zhang}
\email{cyril@mail.ustc.edu.cn}
\affiliation{%
  \institution{University of Science and Technology of China}
}

\author{Yongdong Zhang}
\email{zhyd73@ustc.edu.cn}
\affiliation{%
  \institution{University of Science and Technology of China}
}

\renewcommand{\shortauthors}{Zixiao Wang et al.}

\begin{abstract}
In this paper, we explore the potential of the Contrastive Language-Image Pretraining (CLIP) model in scene text recognition (STR), and establish a novel Symmetrical Linguistic Feature Distillation framework (named CLIP-OCR) to leverage both visual and linguistic knowledge in CLIP. 
Different from previous CLIP-based methods mainly considering feature generalization on visual encoding, 
we propose a symmetrical distillation strategy (SDS) that further captures the linguistic knowledge in the CLIP text encoder. 
By cascading the CLIP image encoder with the reversed CLIP text encoder, a symmetrical structure is built with an image-to-text feature flow that covers not only visual but also linguistic information for distillation.
Benefiting from the natural alignment in CLIP, such guidance flow provides a progressive optimization objective from vision to language, which can supervise the STR feature forwarding process layer-by-layer.
Besides, a new Linguistic Consistency Loss (LCL) is proposed to enhance the linguistic capability by considering second-order statistics during the optimization. 
Overall, CLIP-OCR is the first to design a smooth transition between image and text for the STR task.
Extensive experiments demonstrate the effectiveness of CLIP-OCR with 93.8\% average accuracy on six popular STR benchmarks.
Code will be available at https://github.com/wzx99/CLIPOCR.
\end{abstract}

\ccsdesc[500]{Applied computing~Optical character recognition}

\keywords{scene text recognition;clip;knowledge distillation}


\maketitle

\section{INTRODUCTION}
Scene text recognition (STR) is an important task in Optical Character Recognition (OCR) which aims at reading the text content of the given scene image.
With the widely used of deep learning \cite{fang2022abinet++,niu2020single,ren2019low}, deep neural networks have achieved tremendous improvement on STR task.
But the severe distraction in the image (e.g., noisy background, blurred text, and special text styles) makes it still challenging for recognition by only utilizing the visual information from the input image.

\begin{figure}[!t]
  \centering
  \includegraphics[width=\linewidth]{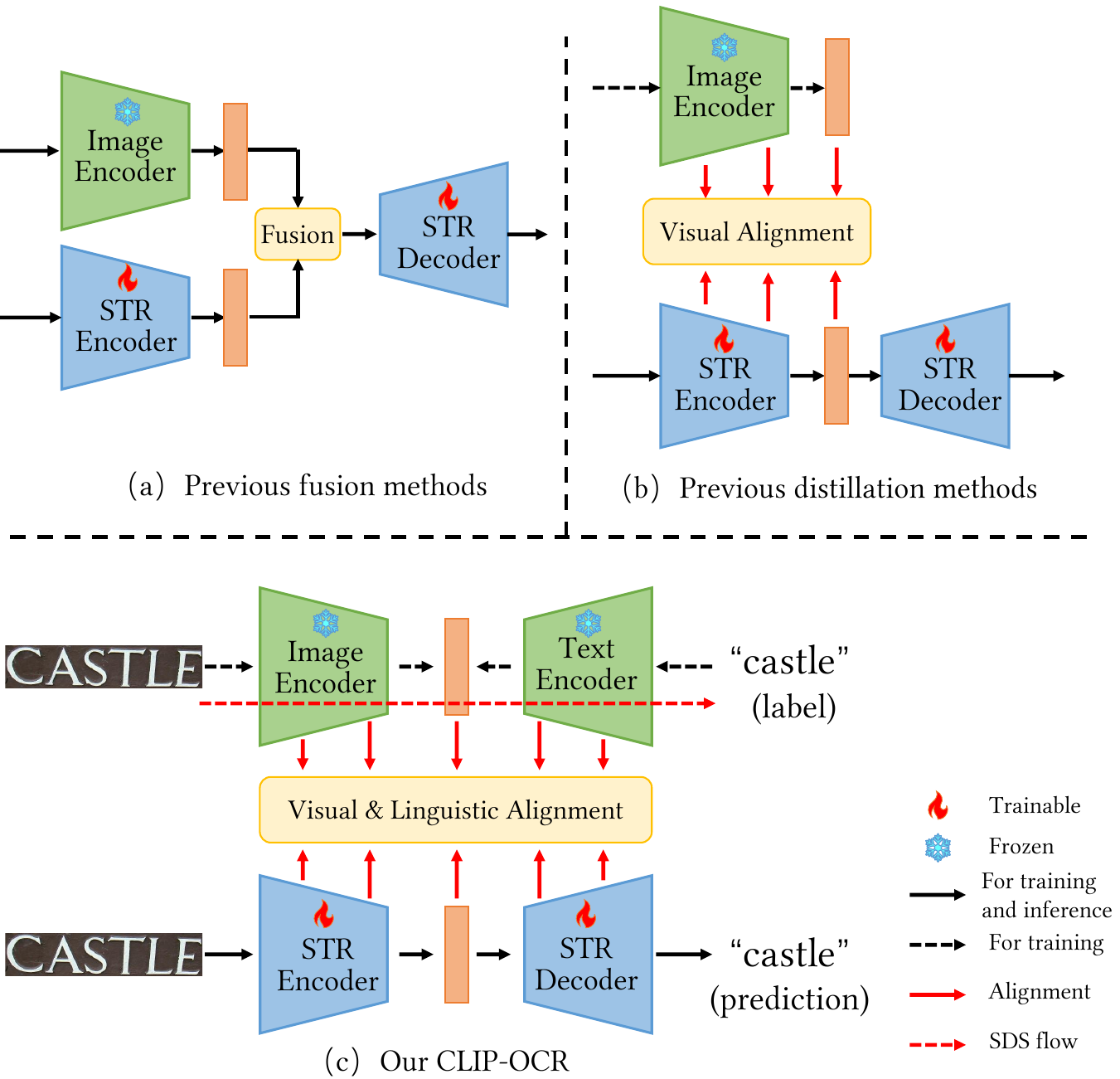}
  \caption{Comparison between previous methods and our CLIP-OCR in STR task. (a) shows the fusion methods which combine the pre-trained CLIP model into their framework to enhance the feature encoder. (b) shows the distillation methods where only visual generalization is concerned for optimizing the feature encoder. (c) is our CLIP-OCR where an SDS flow bridges CLIP encoders together to distill both visual and linguistic knowledge for the entire recognition model.}
  \Description{Difference between previous methods and our CLIP-OCR.}
  \label{img_diff}
\end{figure}

Recently, Contrastive Language-Image Pretraining (CLIP) has shown impressive performance in aligning vision and language content \cite{radford2021learning}.
By leveraging more than 400 million image-text pairs for training, features from CLIP have a powerful generalization property and can be used for few-shot or zero-shot tasks.
Intuitively, since STR involves both visual and language information, CLIP holds promising potential for enhancing recognition performance with its multi-modal prior knowledge.

As a large-scale pre-trained model, CLIP has been widely studied on downstream tasks.
Existing CLIP-based works can be divided into two commonly used methods: fusion \cite{song2022clip, merullo2022linearly,luddecke2022image, ding2022open,yu2023turning} and distillation methods \cite{zhong2022regionclip,xue2022stare,ramesh2022hierarchical, bangalath2022bridging}.
On one hand, fusion methods regard CLIP as a pre-trained feature extractor and directly employ it for feature extraction (Fig.\ref{img_diff}(a)).
For example, TCM \cite{yu2023turning} uses CLIP embeddings to query the text regions inside the image, which verifies the generalizability of CLIP for OCR tasks.
CLIPTER \cite{aberdam2023clipter} introduces the CLIP image encoder to obtain the global visual context and fuse it with the feature of the recognition encoder.
But inferencing with such a large-scale model leads to a huge additional computation cost which is suboptimal for STR task.
On the other hand, distillation methods prefer to transfer knowledge inside CLIP to their model (Fig.\ref{img_diff}(b)).
RegionCLIP \cite{zhong2022regionclip} distills its image encoder with the CLIP text encoder to learn generalized visual representations for open-vocabulary object detection.
Ramesh et al \cite{ramesh2022hierarchical} design an auxiliary loss with the CLIP image encoder to constrain the context of image generation.
However, existing methods mainly focus on distilling visual generalization for image embedding which lacks linguistic guidance when applying to the STR task.
Moreover, the knowledge inside the CLIP text encoder is wasted as there is no text encoding process in STR.
Therefore, a distillation framework for linguistic learning is needed for STR to make full use of the prior knowledge in CLIP. 
In contrast to previous works, in this paper, we further explore the linguistic knowledge in CLIP and propose a novel Symmetrical Linguistic Feature Distillation framework (named CLIP-OCR).

Aiming to incorporate linguistic learning, our CLIP-OCR consists of a Symmetrical Distillation Strategy (SDS) to provide detailed linguistic information and a Linguistic Consistency Loss (LCL) to transition linguistic knowledge efficiently.
Specifically,
the SDS combines CLIP image and text encoder together to generate a precise image-to-text supervision with both visual and linguistic knowledge (Figure 1(c)).
To this end, SDS first introduces the CLIP text encoder and utilizes its linguistic knowledge to guide the recognition decoder.
By exploiting the symmetric input-output relationship between them, we reverse the direction of the CLIP text encoder to create a decoding feature flow.
This reversed feature flow fills the blank of CLIP-based decoder distillation, which creates innovative and seamless linguistic guidance from the text encoder to supervise the recognition decoder.
Second, an image-to-text guidance flow is built to distill the entire recognition process by cascading the CLIP image encoder and reversed text encoder (Red dashed line in Fig.\ref{img_diff}(c)).
Exploiting the well-aligned relations between two CLIP encoders, SDS conducts a progressive layer-wise supervision from image to text that leverages both visual and linguistic knowledge in CLIP.  

In addition, a Linguistic Consistency Loss (LCL) is designed to enhance the transition efficiency of linguistic knowledge.
Different from the general distillation loss with first-order statistics (e.g., point-wise consistency \cite{wei2022contrastive,xue2022stare,heo2019comprehensive}), 
LCL optimizes the recognition model by aligning the second-order statistics to emphasize the learning of inter and intra feature relationship.
With the combining of SDS and LCL, linguistic knowledge in CLIP can be fully transferred to the recognition model to further improve the STR performance.

Overall, by modifying CLIP to a symmetrical recognition flow with visual and linguistic information, CLIP-OCR first builds an instructive and progressive bridge from image to text features in STR, which provides a novel insight for further exploration of CLIP models.
To verify the effectiveness of the proposed method, we evaluate our CLIP-OCR on several mainstream STR benchmarks.
Comprehensive experimental results show the effectiveness of our method with 93.8\% average accuracy which outperforms existing STR methods.

Our main contributions can be summarized as follows:
\begin{itemize}
\item A novel distillation framework (CLIP-OCR) is proposed to provide both visual and linguistic knowledge for STR.
By combining CLIP image and text encoder, CLIP-OCR first conducts a progressive image-to-text flow to guide the recognition layer-wise.
\item We proposed a new Linguistic Consistency Loss (LCL) to enhance the linguistic knowledge capability, which adopts second-order statistics to supply guidance on fine-grained relationships.
\item Experimental results on six STR benchmarks verify the effectiveness of our framework which achieves state-of-the-art performance (93.8\% average accuracy).
\end{itemize}

\section{RELATED WORK}

\begin{figure*}[!t] 
  \centering
  \includegraphics[width=0.9\textwidth]{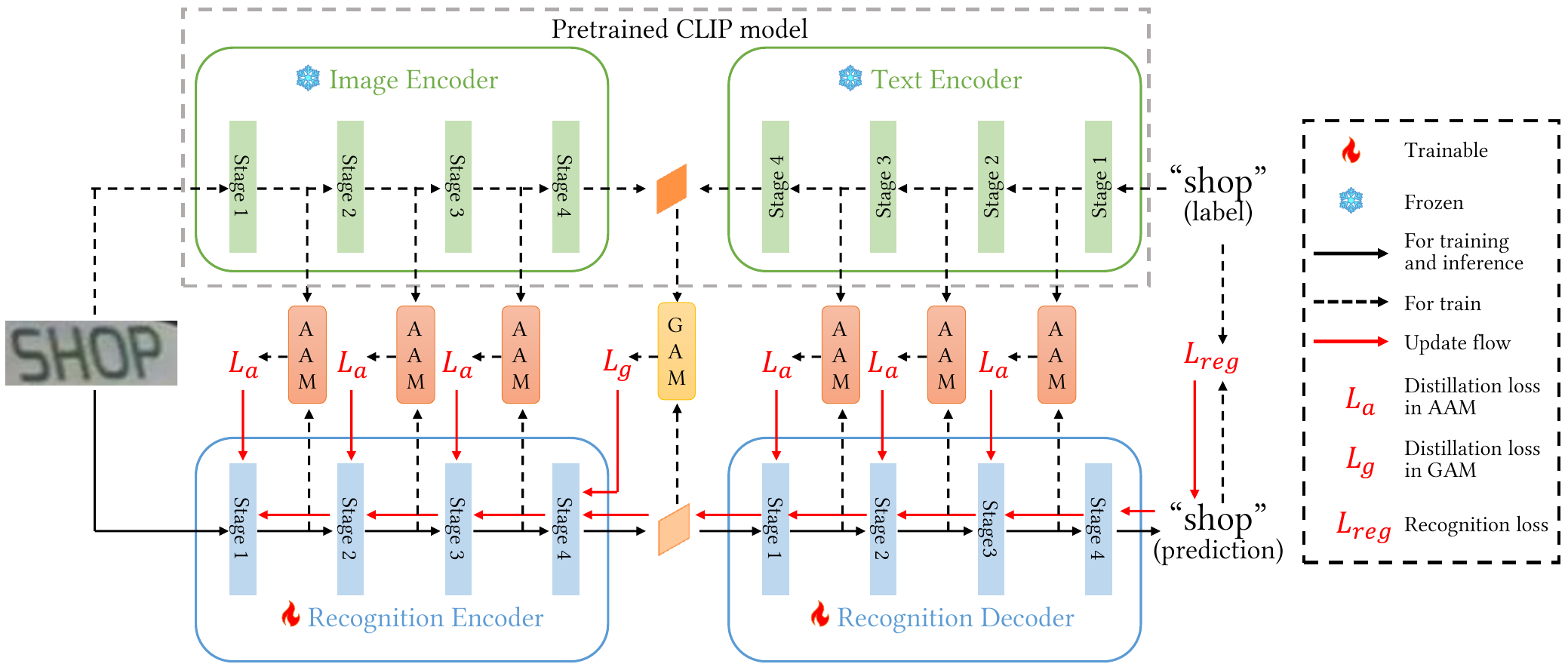}
  \caption{The overall framework of our CLIP-OCR, which consists of a recognition model, CLIP image and text encoder, Adaptive Alignment Module (AAM), and Global Alignment Module (GAM). During training, we first collect the feature maps of each stage from the recognition and CLIP model. Then each feature map from the recognition model is matched with a CLIP feature to calculate the distillation loss La/Lg using AAM/GAM. Finally, the recognition model is optimized by minimizing the combination of distillation loss and general recognition loss.}
  \Description{Framework of CLIP-OCR.}
  \label{img_framework}
\end{figure*}

\subsection{Scene Text Recognition}
Scene text recognition has become a research hotspot with the development of deep learning.
In this section, we divide the existing methods into two categories due to whether used linguistic knowledge: language-free methods and language-aware methods.

For language-free methods, they directly predict the text based on the visual information \cite{shi2016end, wan20192d, baek2019wrong, liao2019scene}.
CRNN \cite{shi2016end} employs VGG \cite{simonyan2014very} for visual feature extracting and RNN for sequence modeling.  
And it introduces Connectionist Temporal Classification (CTC) \cite{graves2006connectionist} for training and prediction.
Compared with CTC, using the attention based head can obtain more accurate performance.
TRBA \cite{baek2019wrong} develops a CRNN framework with thin-plate spline (TPS) transformation and replaces CTC with an attention head.  
ViTSTR \cite{atienza2021vision} designs a simple one-stage framework with Vision Transformer which shows promising results for ViT to apply on STR.

For language-aware methods, they consider the linguistic knowledge inside the word as the auxiliary information to improve the prediction results \cite{fang2021read, wang2020decoupled, wang2021two, lyu2022maskocr, wang2022petr, zhang2023linguistic}.
ABINet \cite{fang2021read} defines a language model with a transformer to refine the results iteratively. 
DAN \cite{wang2020decoupled} decouples the alignment and decoding and uses an RNN-based language model for auto-regression prediction.
Recently, many works prefer to define an auxiliary task to learn linguistic knowledge.
VisionLAN \cite{wang2021two} combines visual and linguistic capability into a single vision model with the help of weakly-supervised masks.
Besides, MaskOCR \cite{lyu2022maskocr} presents a pertain method with MIM to boost the recognition performance.
Previous methods mainly focus on introducing linguistic knowledge by end-to-end auxiliary loss which lack of fine-grained supervision on feature level.
In contrast to existing works, we propose a feature distillation framework with CLIP to provide layer-wise visual and linguistic guidance.

\subsection{Vision-Language Contrastive Learning}
Contrastive Language-Image Pretraining (CLIP) \cite{radford2021learning} designs a multimodal alignment framework with contrastive learning \cite{oord2018representation,li2022deep} to bridge the visual and language information.
It contains an image encoder and a text encoder to measure the content similarity of the given image and text.
Many recent works have transferred it on multiple downstream tasks, including semantic segmentation \cite{luddecke2022image,ding2022open}, object detection \cite{bangalath2022bridging}, Visual Question Answering \cite{wang2022clip} and image generation \cite{ramesh2022hierarchical}.
Many researchers regard CLIP as a pre-trained feature extractor \cite{song2022clip, merullo2022linearly,luddecke2022image, ding2022open,yu2023turning}.
\cite{song2022clip, merullo2022linearly} directly utilize CLIP as an image encoder to extract visual context.
\cite{luddecke2022image, ding2022open} employ CLIP to align the image with the target class for open-vocabulary tasks.
\cite{yu2023turning} uses CLIP embeddings to query the text regions inside the image, which verifies the generalizability of CLIP for OCR tasks.

\subsection{Knowledge Distillation}
Knowledge distillation aims to transfer knowledge from a large-scale model (teacher model) to a lightweight model (student model).
General knowledge distillation methods \cite{hinton2015distilling,li2022dual} use the output of the teacher model to generate soft guidance and force the student model to predict similar results.
With the development of large-scale pre-trained models, many works utilize pre-trained models to distill their model.
Since there is no decoder head in most pre-trained models, \cite{wei2022contrastive} proposes feature distillation to guide the output feature map of the student model.
To measure the similarity in feature level, it designs an additional projection head to embed the features into the same projection space.
Some recent works also combine distillation with CLIP model \cite{zhong2022regionclip,xue2022stare,ramesh2022hierarchical, bangalath2022bridging},
\cite{xue2022stare} distills their backbone with CLIP image encoder and employ Masked Image Modeling (MIM) strategy during training.
\cite{ramesh2022hierarchical, bangalath2022bridging} introduce CLIP as a teacher model to enhance the generalization ability of their model.
In contrast to them, we enable the encoder to distill the decoder layer-wise by further analyzing the relationship between the CLIP text encoder and recognition decoder which expands the scope of applications for distillation with pre-trained models.

\section{METHODOLOGY}
\subsection{Pipeline} 

The overall framework of our CLIP-OCR is illustrated in Fig.\ref{img_framework}.
We follow the teacher-student learning framework \cite{wang2021knowledge} and introduce the CLIP model as the teacher to help the optimization of the student (recognition model).
For the given image, we first use the recognition model to get the predictions and save the recognition feature list.
Meanwhile, the image and its label are also fed into the CLIP image and text encoder, respectively, to generate visual and linguistic guidance.
Especially, since CLIP uses word-level tokenizing which lacks character-level guidance, we split the word-level label into the character-level list for tokenizing to obtain fine-grained feature sequences for feature distillation, e.g., `analysis' $\rightarrow$ `a n a l y s i s' (Section \ref{sec:visual} has verified the alignment ability for the character-level input).
Following the symmetrical distillation strategy (SDS), both the encoder and decoder features from the recognition model can be aligned to the corresponding guidance feature from CLIP.
Then we send the matched feature pairs into the proposed Adaptive Alignment Module (AAM) and Global Alignment Module (GAM) to calculate the Linguistic Consistency Loss (LCL).
And the final loss is obtained by the combination of the distillation loss and the original recognition loss.

\subsection{Symmetrical Distillation Strategy}

\begin{figure}[!t] 
  \centering
  \includegraphics[width=\linewidth]{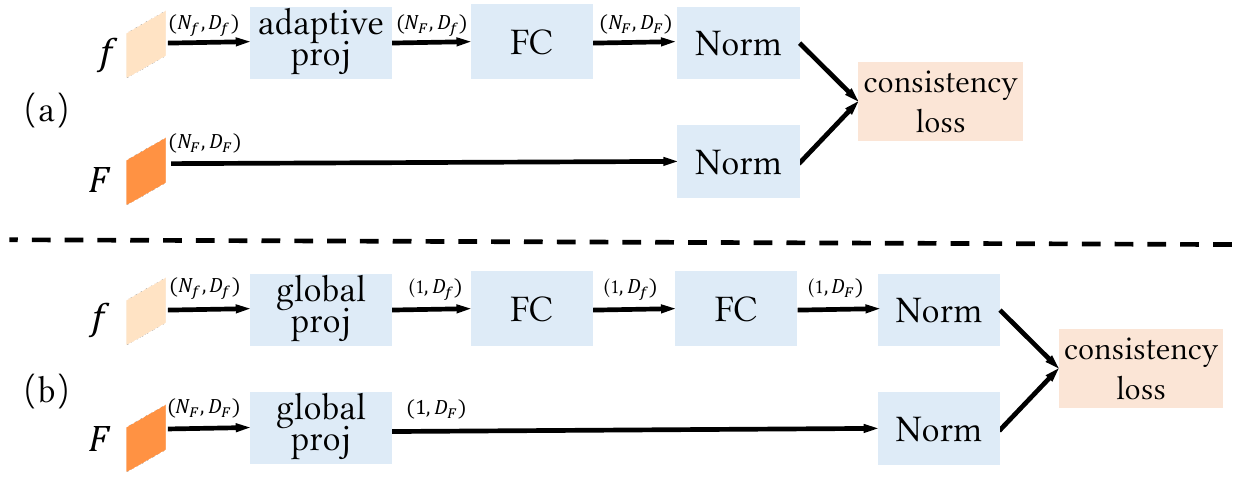}
  \caption{(a) the structure of AAM. (b) the structure of GAM. $(N,D)$ denotes the shape of the feature sequence.}
  \Description{structure of AAM and GAM}
  \label{img_aam_gam}
\end{figure}

Existing distillation methods assume that the input and output data between the teacher and student model should be of the same form for employing consistency supervision.
However, the CLIP model only contains two encoders without the decoder which encounters the structure mismatching problem on distilling encoder-decoder models. 
For most recognition models with encoder-decoder structure, although distilling features layer-wise allows the CLIP image encoder to guide the recognition encoder(Fig.\ref{img_diff}(b)), such a simple distillation strategy not only lacks supervision on the decoder but also wastes the linguistic knowledge inside the CLIP text encoder, which is fatal for STR models.

To address this issue, we propose a symmetrical distillation strategy (SDS) to provide visual as well linguistic knowledge for the whole recognition process.
The key idea of SDS is that we introduce linguistic information in the CLIP text encoder by leveraging it to distill the recognition decoder.
As the input and output of them are symmetrically related, we let the deep layers of the recognition decoder be aligned with the shallow layer of the CLIP text encoder.
This is reasonable as decoding forward is a reconstruction process that maps from a complex, high-dimensional feature space to a simpler text embedding space.
Thus, a decoding stream from feature to text can be built by reversing the feature flow of the CLIP text encoder.
After cascading with the visual encoding stream in the CLIP image encoder, as shown in Fig.\ref{img_framework}, the CLIP encoders produce a symmetrical image-to-text flow to make full use of their visual and linguistic knowledge.
If we regard such image-to-text flow as an encoder-decoder stream for recognition, instructive layer-wise guidance can be created to optimize recognition models seamlessly.

Since directly aligning the student model with the teacher model may affect the generalization of student feature space \cite{gupta2022understanding}, we propose the Adaptive Alignment Module (AAM) and the Global Alignment Module (GAM) for feature projection.
Assuming the recognition feature $f$ with the shape of $(N_f, D_f)$ and CLIP feature $F$ with the shape of $(N_F, D_F)$.
As shown in Fig.\ref{img_aam_gam}, AAM first uses a trainable adaptive projection matrix $P \in (N_F, N_f)$ and a linear layer $W_1 \in (D_f,D_F)$ to project $f$ to the feature space of $F$ and adjust their shapes to be the same.
Then, a normalization layer is used to eliminate the influence of magnitude before calculating the consistency loss.
The distillation loss $L_{a}$ of AAM can be formulated as follows, 

\begin{equation} \label{eq:aam}
L_{a}(f,F) = l(norm(P \times f \times W_1),norm(F)),
\end{equation}
where $l$ is a predefined consistency loss (e.g., L1 loss), $\times$ is the matrix multiplication and $norm$ denotes normalization.
For GAM, it is designed for bridging the guidances progressively from the CLIP image encoder to the text encoder by utilizing their alignment property in the class token.
As Eq.\ref{eq:gam} shows, GAM follows a similar process as AAM but only selects the class token by the global projection.

\begin{equation} \label{eq:gam}
L_{g}(f,F) = l(norm(g(f^{cls} \times W_1) \times W_2),norm(F^{cls})),
\end{equation}
where $f^{cls}$ and $F^{cls}$ denote the class token from recognition and CLIP model, $W_1, W_2$ denote linear layers, and $g$ is the activate layer (ReLU) .

By applying AAM and GAM on the image-to-text flow, SDS generates the layer-wise supervision as follows,

\begin{equation} \label{eq:sds}
L_{dis} = \sum^{3}_{i=1} L_{a}(f^{E}_{i}, F^{I}_{i}) + L_{g}(f^{E}_{4}, F^{T}_{4})  + \sum^{3}_{i=1} L_{a}(f^{D}_{i}, F^{T}_{4-i}),
\end{equation}
where, $f^{E},f^{D}$ are the features from recognition encoder and decoder, $F^{I},F^{T}$ are the features from CLIP image and text encoder.
$i$ is the stage index of the feature sequence where we spilt all parts of models into 4 stages as shown in Fig.\ref{img_framework}.
In Eq.\ref{eq:sds}, the first and third terms indicate the learning of visual and linguistic knowledge.
Since there is a strong alignment between CLIP image and text encoder, the second term of Eq.\ref{eq:sds} becomes a bridge between the visual to linguistic knowledge which guide the feature stream progressively.
Overall, both encoders of CLIP are utilized in SDS for supplying comprehensive visual and linguistic expertise that covers the entire recognition pipeline.

\subsection{Linguistic Consistency Loss}
General distillation loss primarily considers the consistency of the first-order statistics, such as point-wise L1 loss in Eq.\ref{eq:l1}.

\begin{equation} \label{eq:l1}
l_{L1}(f,F) = \frac{1}{ND}||f - F||_1,
\end{equation}
where $f,F \in (N,D)$ are the feature sequences from AAM or GAM.
However, these basic learning objectives can not well transfer the linguistic knowledge inside the relationship between characters, which is important for language modeling in the STR task.
To enhance the efficiency of learning linguistic knowledge from SDS, we propose a Linguistic Consistency Loss (LCL) to develop linguistic capability by emphasizing the alignment of intra and inter relationship.

In LCL, the relationship is measured by second-order statistics: self-attention map and cross-attention map.
For intra-relationship alignment, we measure the character-level similarity by self-attention map from recognition and CLIP model.
And their intra consistency loss can be formulated as follows,

\begin{equation} \label{eq:intralcl}
l_{intra}(f,F) = \frac{1}{N^2}||f \times f^T - F \times F^T||_1,
\end{equation}

For inter-relationship alignment, we define a contrastive learning loss between recognition and CLIP features to enhance the discrimination of feature sequences and reduce ambiguous linguistic information.
As shown in Eq.\ref{eq:interlcl}, the cross-attention map between recognition and CLIP features is utilized as their similarity and we calculate cross entropy loss with a small temperature factor $\tau$.

\begin{equation} \label{eq:interlcl}
l_{inter}(f,F) = \frac{1}{N}\mathrm{cross\_entropy}(f \times F^T/\tau, I),
\end{equation}
where $I$ is the diagonal matrix which serves as the label for $l_{inter}$, $\tau$ is set to 0.03 to highlight the difference between the features.

By combining the $l_{intra}$ and $l_{inter}$, the LCL is calculated by the weighted sum of them,

\begin{equation} \label{eq:lcl}
l_{LCL}(f,F) = \lambda_1 l_{intra}(f,F) + \lambda_2 l_{inter}(f,F),
\end{equation}
where the $\lambda_1$ and $\lambda_2$ are the super-parameters determined experimentally.

Finally, we apply our LCL on Eq.\ref{eq:aam}-\ref{eq:gam} and the total loss function is formulated in Eq.\ref{eq:fin}.
For inference time, we remove the CLIP model, AAM, and GAM where only original recognition is needed.

\begin{equation} \label{eq:fin}
L = L_{reg} + L_{dis},
\end{equation}
where $L_{reg}$ denotes the regular recognition loss (e.g., character-level cross entropy loss).

\section{EXPERIMENTS} 


\subsection{Datasets} \label{sec:data}

Following \cite{fang2021read,bautista2022scene}, we use two synthetic datasets (MJ \cite{jaderberg2014synthetic} and ST \cite{gupta2016synthetic}) for training and evaluate our method on six standard datasets (IIIT \cite{mishra2012scene}, SVT \cite{wang2011end}, IC13 \cite{karatzas2013icdar}, IC15 \cite{karatzas2015icdar}, SVTP \cite{phan2013recognizing}, CT \cite{risnumawan2014robust}). 
Following \cite{wang2022multi}, we split the benchmarks into regular text benchmarks (IIIT, SVT, IC13) and irregular text benchmarks (IC15, SVTP, CT).
Moreover, we also introduce five additional challenging datasets for further evaluation, including ArT \cite{chng2019icdar2019}, COCO-Text (COCO) \cite{veit2016coco}, Uber-Text (Uber) \cite{zhang2017uber}, WordArt \cite{xie2022toward}, and OST \cite{wang2021two}.

\subsection{Implementation Details} \label{sec:imp}

For training settings,
We use the Adam optimizer with a learning rate of 1.4e-3.
We set the batch size to 320 and train the network for 5 epochs.
Images are resized to 32 $\times$ 128.
Following \cite{bautista2022scene},  the RandAugment is utilized for data augmentation, including Sharpness, Invert, GaussianBlur, and PoissonNoise.
As for super-parameters in our method, we experimentally set $\tau$ to 0.03, $\lambda_1$ to 5, and $\lambda_1$ to 0.1.
The experiments are employed with the PyTorch framework on 2 NVIDIA RTX 2080ti GPUs.

For the CLIP model, we use the CLIP with ViT-B/16 as the teacher model in distillation. 
We reduce the number of positional embedding to fit the input size of 32 $\times$ 128.
All parameters in CLIP are frozen during the training process.
Different from using word-level tokenize in the original CLIP, we split the word into a character-level list before sending it to the text encoder.

For the recognition model, we define a simple transformer-based model as the baseline.
We use the ViT-Small as the encoder which contains 12 transformer encoder layers.
The encoder layers are split into 4 stages for distillation where each stage contains 3 transformer encoder layers.
And we introduce 4 transformer decoder layers as the recognition decoder where each decoder layer stands for a stage for distillation.
During the inference stage, we adopt the autoregressive decoding strategy.

\begin{table}
  \caption{The effectiveness of SDS. `Image' means using CLIP image encoder and `Text' means using CLIP text encoder.}
  \label{tab:sds}
  \begin{tabular}{c | c | c | c | c | c | c | c | c }
    \toprule
    Image & Text & IIIT & SVT & IC13 & IC15 & SVTP & CT & Avg\\
    \midrule
    - & -  & 96.6 & 95.5 & 97.5 & 85.9 & 88.8 & 89.6 & 93.0 \\
    \checkmark & - & 96.8 & 94.4 & 97.8 & 87.0 & 90.4 & 92.1 & 93.5 \\
     - & \checkmark & 97.3 & 93.8 & 97.9 & 86.4 & 90.7 & 88.2 & 93.4 \\
    \checkmark & \checkmark  & 97.3 & 94.7 & 97.7 & 87.2 & 89.9 & 93.1 & \textbf{93.8} \\
  \bottomrule
\end{tabular}
\end{table}

\begin{table}
  \caption{The effectiveness of LCL.}
  \label{tab:lcl}
  \begin{tabular}{c | c | c | c | c | c | c | c | c }
    \toprule
    $l_{intra}$ & $l_{inter}$  & IIIT & SVT & IC13 & IC15 & SVTP & CT & Avg\\
    \midrule
    - & - & 97.1 & 95.1 & 97.6 & 86.4 & 89.6 & 91.3 & 93.4 \\
    \checkmark & - & 97.2 & 94.9 & 97.4 & 87.0 & 89.9 & 92.0 & 93.6\\
     - & \checkmark & 96.9 & 95.1 & 97.8 & 86.9 & 89.9 & 91.7 & 93.5 \\
    \checkmark & \checkmark & 97.3 & 94.7 & 97.7 & 87.2 & 89.9 & 93.1 & \textbf{93.8} \\
  \bottomrule
\end{tabular}
\end{table}

\begin{table}
  \caption{Comparision with other distillation loss.}
  \label{tab:distill_loss}
  \begin{tabular}{c | c | c | c | c | c | c | c }
    \toprule
    Method  & IIIT & SVT & IC13 & IC15 & SVTP & CT & Avg\\
    \midrule
    - & 96.0 & 95.5 & 97.5 & 85.9 & 88.8 & 89.6 & 93.0 \\
    L1 & 97.1 & 95.1 & 97.6 & 86.4 & 89.6 & 91.3 & 93.4 \\
    cos & 96.8 & 94.3 & 97.1 & 87.2 & 89.0 & 91.7 & 93.3 \\
    KL & 96.9 & 94.7 & 97.3 & 86.6 & 89.8 & 90.3 & 93.3 \\
    LCL & 97.3 & 94.7 & 97.7 & 87.2 & 89.9 & 93.1 & \textbf{93.8} \\
  \bottomrule
\end{tabular}
\end{table}

\subsection{Evaluation Metric} \label{sec:metric}
We set the size of the recognition character set to 36, including a-z and 0-9.
The word accuracy is used as the evaluation metric, where a correct word means all characters should be matched to the label.
Following \cite{baek2021if}, we further report the weighted average score (Avg) based on the sample number in each dataset. 

\subsection{Ablation Study} \label{sec:ablation}

\noindent\textbf{The effectiveness of SDS:} 
First, we study the effectiveness of each component in SDS.
As shown in Table \ref{tab:sds}, our baseline is 93.0\% average accuracy, and distillation with CLIP image encoder can obtain 93.5\%.
After adding the CLIP text encoder with SDS, we further improve the performance to 93.8\%.
This result demonstrates that the visual and linguistic knowledge provided by the CLIP image and text encoder is complementary to each other.
When only using CLIP text encoder to guide the reignition decoder, we also can get 93.4\% average accuracy.
This again shows that our SDS can utilize the linguistic knowledge in the CLIP text encoder to enhance the STR performance.

\noindent\textbf{The effectiveness of LCL:} 
To evaluate the LCL, we compare the results with $l_{inter}$ or $l_{intra}$ in Table \ref{tab:lcl}.
The first line in Table \ref{tab:lcl} is the baseline model distilled with SDS and point-wise L1 loss.
It can be seen that the combining of $l_{inter}$ and $l_{intra}$ can improve the performance from 93.4\% to 93.8\%.
When only using $l_{intra}$ or $l_{inter}$, we still get 0.2\% and 0.1\% improvements.
This justifies that our distillation framework can transfer more knowledge from CLIP to the recognition model.
And the 0.4\% total improvements from LCL show the two parts in our loss have well cooperated during the training process.
Compared with $l_{inter}$, $l_{intra}$ performs slightly better which shows the importance of capturing intra-relationship in STR task. 
Moreover, Table \ref{tab:distill_loss} compares our LCL with other general distillation loss, including point-wise L1 loss, cosine similarity, and KL-divergence loss.
Results have shown that our LCL with second-order statistics outperforms other methods with first-order statistics.
This verifies that our method can generate more suitable guidance for the STR task.

\begin{table}
  \caption{The influence of $\tau$.}
  \renewcommand{\arraystretch}{1.0}
  {
  \begin{tabular}{c | c | c | c | c }
    \toprule
     $\tau$ &  0.01 & 0.03 & 0.05 & 1\\
    \midrule
    Avg & 93.6 & \textbf{93.8} & 93.7 & 93.4\\
  \bottomrule
  \end{tabular}
  }
\label{tab:tau}
\end{table}

\begin{table}
  \caption{The influence of $\lambda_1$. $\lambda_2$ is set to 0.1 by default.}
  \renewcommand{\arraystretch}{1.0}
  {  
  \begin{tabular}{c | c | c | c | c | c }
    \toprule
     $\lambda_1$ & 1 & 3 & 5 & 7 & 10 \\
    \midrule
    Avg & 93.6 & 93.6 & \textbf{93.8} & 93.6 & 93.4 \\
  \bottomrule
\end{tabular}  
  }
\label{tab:lambda_1}
\end{table}

\begin{table}
  \caption{The influence of $\lambda_2$. $\lambda_1$ is set to 5 by default.}
  \renewcommand{\arraystretch}{1.0}
  {  
  \begin{tabular}{c | c | c | c | c }
    \toprule
     $\lambda_2$ &  0.01 & 0.1 & 0.3 & 1 \\
    \midrule
    Avg & 93.6 & \textbf{93.8} & 93.7 & 93.1 \\
  \bottomrule
\end{tabular}
}
\label{tab:lambda_2}
\end{table}

\begin{table}
  \caption{The influence of CLIP model.}
  \label{tab:clip}
  \begin{tabular}{c | c | c | c }
    \toprule
     Model & Token & CLIP & Avg\\
    \midrule
    ViT-S & char & CLIP-ViT-B/32 & 93.6 \\
    ViT-S & char & CLIP-ViT-B/16 & 93.8 \\
    ViT-S & word & CLIP-ViT-B/16 & 93.4 \\
    CLIP-ViT-B/16 & - & CLIP-ViT-B/16 & 81.1 \\
  \bottomrule
\end{tabular}
\end{table}

\begin{table}
  \caption{Language Capability on OST dataset.}
  \label{tab:ost}
  \begin{tabular}{c | c | c | c}
    \toprule
     Method &  Weak & Heavy & Avg\\
    \midrule
    RNN & 63.9 & 43.9 & 53.9 \\
    Transformer & 68.4 & 48.0 & 58.2 \\
    VisionLAN & 70.3 & 50.3 & 60.3 \\
    CLIP-OCR & 80.5 & 69.7 & \textbf{75.1} \\
  \bottomrule
\end{tabular}
\end{table}

\begin{table*}
  \caption{Comparison with SOTA methods on six STR benchmarks. Bold and underlined values denote the 1st and 2nd results in each column.}
  \renewcommand{\arraystretch}{1.0}
  {
  \begin{tabular}{l|c|c|c|c|c|c|c|c|c}
    \toprule
     \multirow{2}{*}{Methods} & \multirow{2}{*}{Language} &  \multicolumn{3}{c|}{Regular Text} & \multicolumn{3}{c|}{Irregular Text} & \multirow{2}{*}{Avg}& \multirow{2}{*}{Params(M)}\\
    \cline{3-8}
     & & IIIT & SVT& IC13 & IC15 & SVTP & CT & \\
    \midrule
    CRNN \cite{shi2016end} & $\times$ & 82.9 & 81.6 & 91.9 & 69.4 & 70.0 & 65.5 & 78.6 & 8.3\\
    TRBA \cite{baek2019wrong} & $\times$ & 87.9 & 87.5 & 93.6 & 77.6 & 79.2 & 74.0 & 84.6 & -\\
    DAN \cite{wang2020decoupled} & $\times$ & 94.3 & 89.2 & 93.9 & 74.5 & 80.0 & 84.4 & 87.2 & - \\
    RobustScanner \cite{yue2020robustscanner} & $\times$ & 95.3 & 88.1 & 94.8 & 77.1 & 79.5 &  90.3 & 88.4 & - \\
    TextScanner \cite{wan2020textscanner} & $\times$ & 93.9 & 90.1 & 92.9 & 79.4 & 84.3 & 83.3 & 88.5 & - \\
    ViTSTR \cite{atienza2021vision} & $\times$ & 88.4 & 87.7 & 93.2 & 78.5 & 81.8 & 81.3 & 85.6 & - \\   
    SVTR \cite{du2022svtr} & $\times$ & 96.0 & 91.5 & 97.1 & 85.2 & 89.9 & 91.7 & 92.3 & 24.6 \\
    \midrule
    SEED \cite{qiao2020seed} & \checkmark & 93.8 & 89.6 & 92.8 & 80.0 & 81.4 & 83.6 & 88.3 & -\\
    VisionLAN \cite{wang2021two} & \checkmark & 95.8 & 91.7 & 95.7 & 83.7 & 86.0 & 88.5 & 91.2 & 32.8\\
    ABINet \cite{fang2021read} & \checkmark & 96.2 & 93.5 & 97.4 & 86.0 & 89.3 & 89.2 & 92.3 & 36.7 \\
    Parseq$_A$ \cite{bautista2022scene} & \checkmark & \underline{97.0} & 93.6 & 97.0 & 86.5 & 88.9 & \underline{92.2} & \underline{93.3} & 23.8 \\
    MGP \cite{wang2022multi} & \checkmark & 95.3 & 93.5 & 96.4 & 86.1 & 87.3 & 87.9 & 92.0 & 52.6 \\
    \midrule
    ConCLR \cite{zhang2022context} & \checkmark & 96.5  & 94.3 & \textbf{97.7} & 85.4 & 89.3  &  91.3 & 92.8 & 37.0 \\
    MaskOCR \cite{lyu2022maskocr} & \checkmark & 95.5  & \textbf{95.7} & \underline{97.1} & \underline{87.0} & \textbf{90.1} &  90.3 & 92.9 & 31.0 \\
    \midrule 
    CLIP-OCR(ours) & \checkmark & \textbf{97.3} & \underline{94.7} & \textbf{97.7} & \textbf{87.2} & \underline{89.9} & \textbf{93.1} & \textbf{93.8} & 31.1 \\
    \bottomrule
  \end{tabular}
  }
  \label{tab:sota}
\end{table*}

\begin{table}
  \caption{Comparison with SOTA methods on challenging datasets.}
  \renewcommand{\arraystretch}{1.0}
  {
    \begin{tabular}{c | c | c | c | c}
      \toprule
       Method &  ArT & COCO & Uber & WordArt\\
      \midrule
      CRNN \cite{shi2016end} & 57.3 & 49.3 & 33.1 & 47.5\\
      ViTSTR \cite{atienza2021vision} & 66.1 &  56.4 &  37.6 & -\\
      TRBA \cite{baek2019wrong} & 68.2 &  61.4 &  38.0 & 55.8\\
      ABINet \cite{fang2021read} & 65.4 & 57.1 & 34.9 & 67.4\\
      PARSeq$_A$ \cite{bautista2022scene} & \textbf{70.7} & 64.0 & 42.0 & -\\
      CornerTransformer \cite{xie2022toward} & - & - & - & 70.8 \\
      CLIP-OCR & 70.5 & \textbf{66.5} & \textbf{42.4} & \textbf{73.9}  \\
    \bottomrule
    \end{tabular}
  }
\label{tab:sota-new}
\end{table}

\begin{figure}[t]
  \centering
  \includegraphics[width=\linewidth]{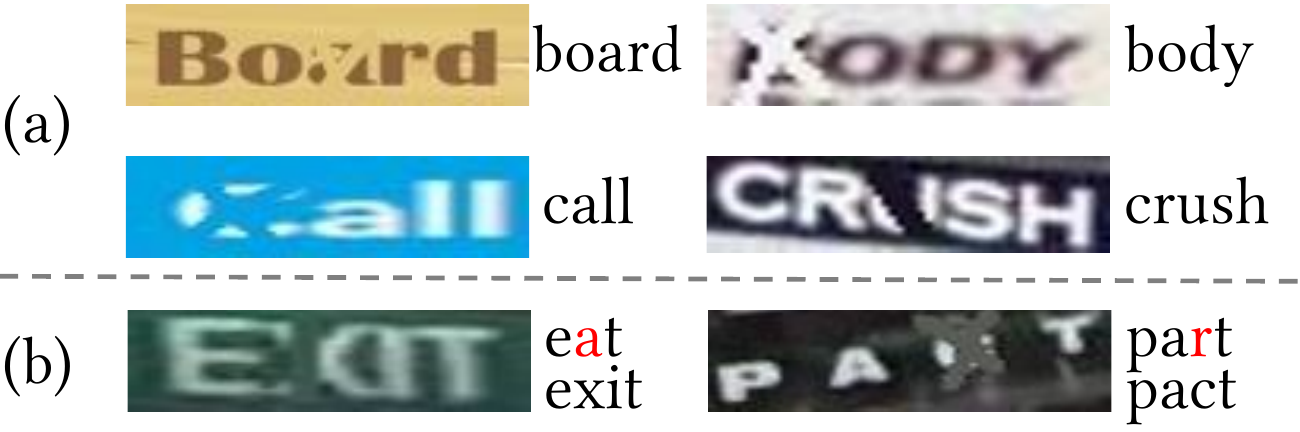}
  \caption{Visualization results on OST. (a) Successfully recognized results. (b) Unsuccessfully recognized results. The first line is prediction and the second line is label. }
  \Description{Visualization results on OST.}
  \label{img_ost}
\end{figure}

\noindent\textbf{The influence of super-parameters in LCL:} 
Table \ref{tab:tau} shows the influence of $\tau$ in Eq.\ref{eq:interlcl}, results show that a smaller $\tau$ is necessary for $l_{inter}$ as it needs to emphasize the discrimination of the feature.
When $\tau<0.05$, we found that it does not have a large impact on the final results, and the best performance is captured by setting $\tau$ to 0.03.

Besides, Table \ref{tab:lambda_1} and Table \ref{tab:lambda_2} illustrate the influence of the $\lambda_1$ and $\lambda_2$ in Eq.\ref{eq:lcl}.
The results show that our method can obtain the best performance with $\lambda_1=5$ and  $\lambda_2=0.1$.
And the results is stable for $\lambda_1 \in [1,7]$ and  $\lambda_2 \in [0.01,0.3]$.
But an extremely large value may affect the optimization with original recognition loss $L_{reg}$, resulting in a performance drop.

\noindent\textbf{The influence of CLIP model:}
We compare our method with different sizes of CLIP in Table \ref{tab:clip}.
The results show that a larger CLIP model leads to better performance which is reasonable in the distillation framework.
In the third line, We stop to split the label into the character-level list and directly use the word-level tokenizing for CLIP text encoder, which results in 0.4\% performance drop and even lower than only using image encoder in Table \ref{eq:sds} (93.5\%).
This is because word-level tokenizing reduces the length of the feature sequence to three (the start token, word token, and end token) for most samples, which fails to provide character-level guidance for recognition and affects the fine-grained capability of the decoder.
In addition, we also compare our CLIP-OCR with directly using the CLIP image encoder as the recognition encoder.
Following \cite{zhou2022learning}, we freeze the CLIP image encoder and use promote learning to reduce the trainable parameter size.
However, Table \ref{tab:clip} shows that simply using the CLIP image encoder as the feature extractor can not provide sufficient fine-grained information for recognition.
Thus, our CLIP-OCR with AAM and GAM to align features in the projection space provides a necessary and effective way for leveraging the knowledge inside CLIP for the STR task.

\subsection{Linguistic Capability on Occluded Data}
To evaluate the linguistic capability of the proposed CLIP-OCR, we evaluate our model on the OST dataset \cite{wang2021two}.
OST dataset contains 4832 manually occluded recognition images which are divided into weak and heavy sets according to the occluded degree.
As shown in Table \ref{tab:ost}, we achieve 75.1\% accuracy which outperforms the previous methods.
Fig.\ref{img_ost} further shows our CLIP-OCR can infer the occluded characters with its linguistic knowledge.
For the extremely occluded images (the third line in Fig.\ref{img_ost}(b)), CLIP-OCR still can predict a reasonable word.
The experiment results on OST have shown that our CLIP-OCR can leverage linguistic information to enhance the language modeling ability of the recognition model.

\subsection{Comparisons with State-of-the-Arts} \label{sec:sota}
We compare our method with multiple recent state-of-the-art recognition methods on 6 benchmarks in Table.\ref{tab:sota}.
All methods are trained in synthetic datasets for comparison. 
Based on whether use the linguistic information, we divide the existing method into language-free methods (CRNN, TRBA, DAN, RobustScanner, RobustScanner, ViTSTR, SVTR) and language-aware methods (SEED, VisionLAN, ABINet, Parseq, MGP).
Besides, we further compare with some recent pre-trained methods (ConCLR and MaskOCR).

As shown in Table.\ref{tab:sota}, generally, the language-aware methods have better performance than language-free methods which demonstrates the importance of linguistic information.
For the average accuracy, our CLIP-OCR obtains 93.8\% which outperforms other methods.
Specifically, our method achieves the 1st performance on IIIT, IC13, IC15, and CT datasets (with 0.3\%, 0.6\%, 0.2\% , and 0.9\% improvements) and the 2nd performance on SVT and SVTP datasets.
Compared with language-aware methods, our CLIP-OCR obtain over 0.5\% improvements in average accuracy. 
For methods with additional word-level supervision (SEED and MGP), our CLIP-OCR surpasses them by over 1.8\% under the comparable model size. 
These results show that our CLIP-OCR with linguistic distillation can generate more effective guidance for language modeling to enhance recognition performance.
Besides, our distillation framework also outperforms the pre-trained method ConCLR and MaskOCR with comparable parameters, which shows that our distillation framework is an effective way to obtain accurate and lightweight STR models.
In Table.\ref{tab:sota-new}, we further evaluate our method on 4 additional challenging datasets: ArT, COCO, Uber, and WordArt. Results show that our method also achieves stat-of-the-art performance which again verifies the effectiveness of our CLIP-OCR.

\subsection{Qualitative Analysis} \label{sec:visual}
\begin{figure}[t]
  \centering
  \includegraphics[width=\linewidth]{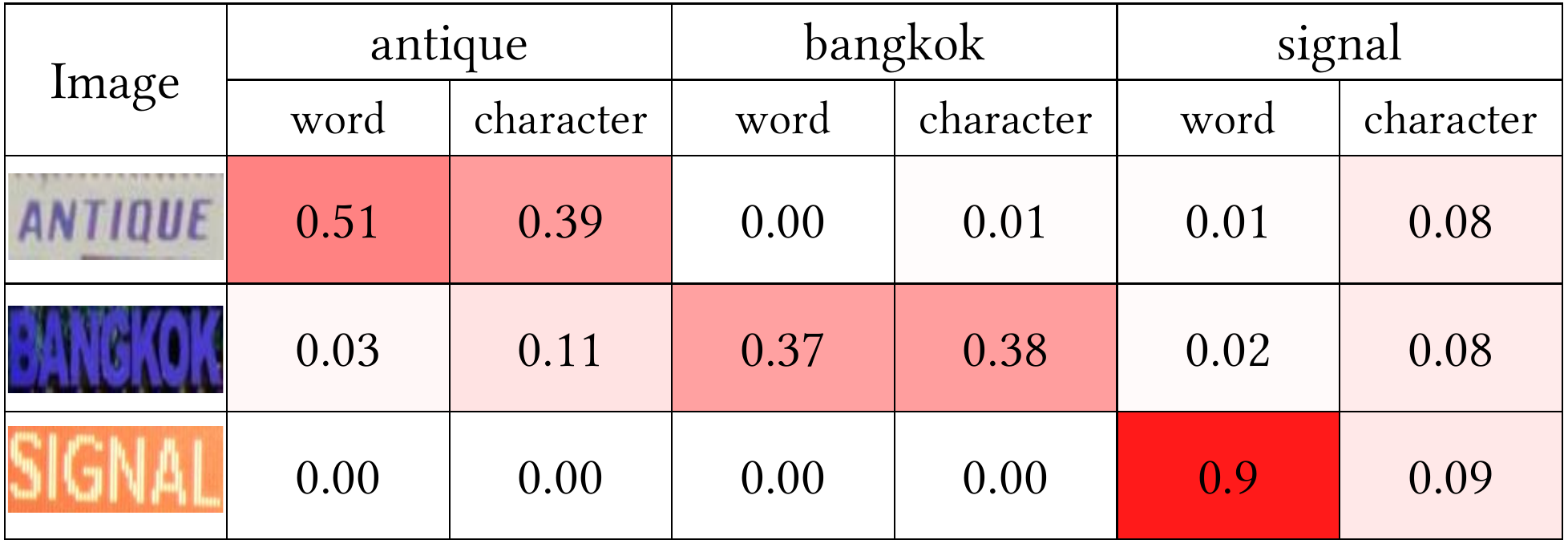}
  \caption{Character-level alignment ability of CLIP model. "word" and "character" means world-level and character-level tokenizing, respectively.}
  \Description{Character-level alignment ability of CLIP model.}
  \label{img_clip}
\end{figure}

\noindent\textbf{Character-level alignment ability of CLIP model:}
Different from the original input format in CLIP which requires 224$\times$224 image size for the image encoder and world-level texts for the text encoder, 
we resize the images to a small size and split the word-level text label into the character-level list as the input.
To verify whether CLIP is capable enough in our framework, we further visualize the alignment ability in Fig.\ref{img_clip}.
Specifically, we first choose 3 recognition images and resize them to 32$\times$128 for the CLIP image encoder. 
Then we send their labels to the CLIP text encoder with both word-level and character-level tokenizing.
The results in Fig.\ref{img_clip} show that the CLIP still has strong alignment ability under the 32$\times$128 input size.
And CLIP is also able to align the character-level input with the corresponding image which provides powerful support for our CLIP-OCR.

\begin{figure}[t]
  \centering
  \includegraphics[width=0.95\linewidth]{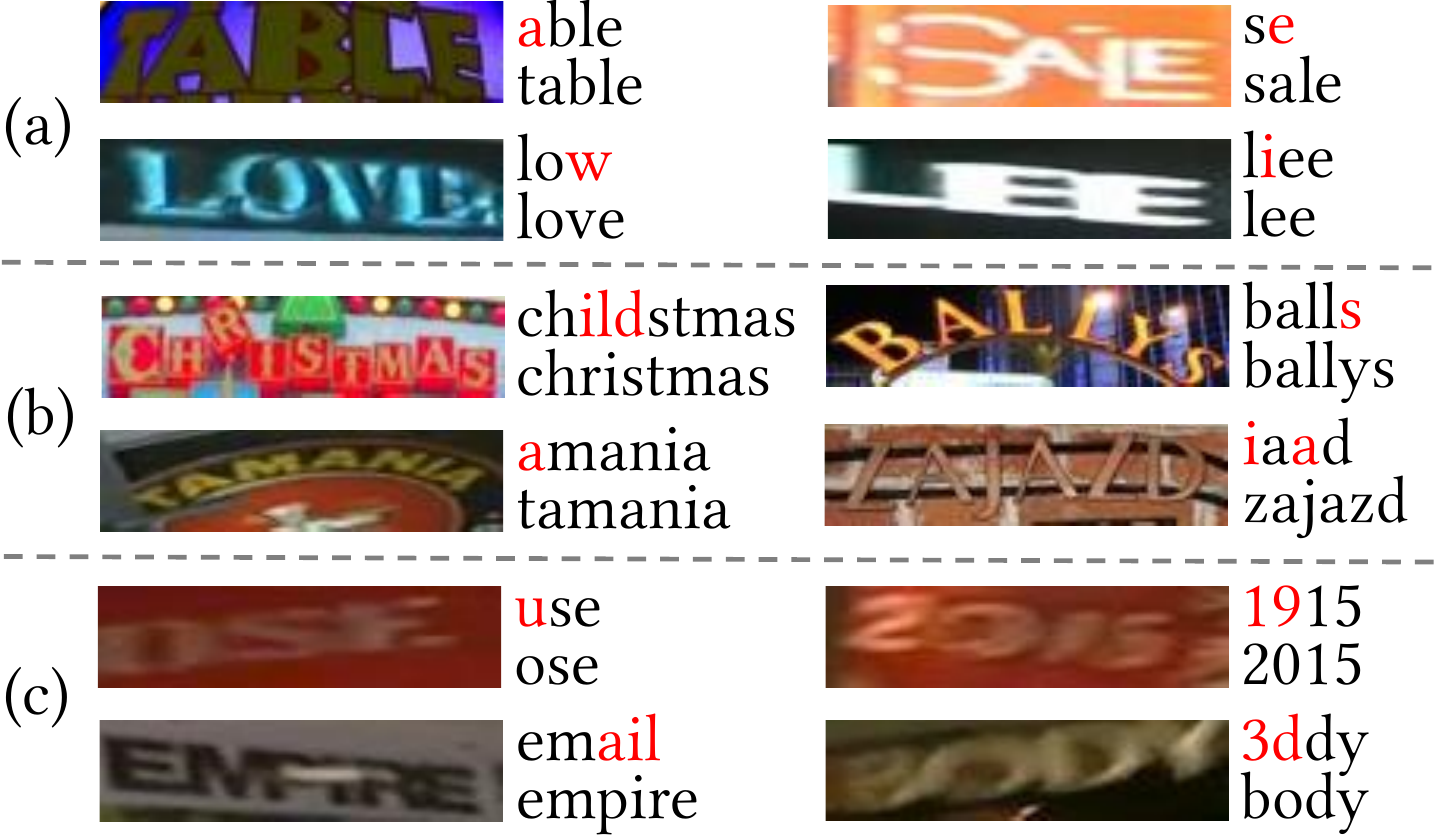}
  \caption{Recognition results of CLIP-OCR. (a) samples with specially shaped texts. (b) samples with complex backgrounds. (c) low-quality samples. The first line is the prediction with the baseline model, and the second line is the result of our CLIP-OCR.}
  \Description{Recognition results of CLIP-OCR.}
  \label{img_vis}
\end{figure}

\noindent\textbf{The effectiveness of CLIP-OCR:}
To verify the effectiveness of CLIP-OCR qualitatively, Fig.\ref{img_vis} collects some recognition results to demonstrate the effect of distilling linguistic knowledge in CLIP-OCR.
Fig.\ref{img_vis}(a) shows our CLIP-OCR can deal with specially shaped texts (e.g., the `t' in `table' and the `al' in `sale').
Fig.\ref{img_vis}(b) presents some cases with a complex background where our CLIP-OCR also can avoid the distraction with non-text regions (e.g., `y' in `ballys', `z' in `zajazd').
And Fig.\ref{img_vis}(c) further shows the effectiveness of our method on images with low quality (e.g., `o' in `ose', `bo' in `body'). 

\subsection{Discussion}

\begin{figure}[t]
  \centering
  \includegraphics[width=\linewidth]{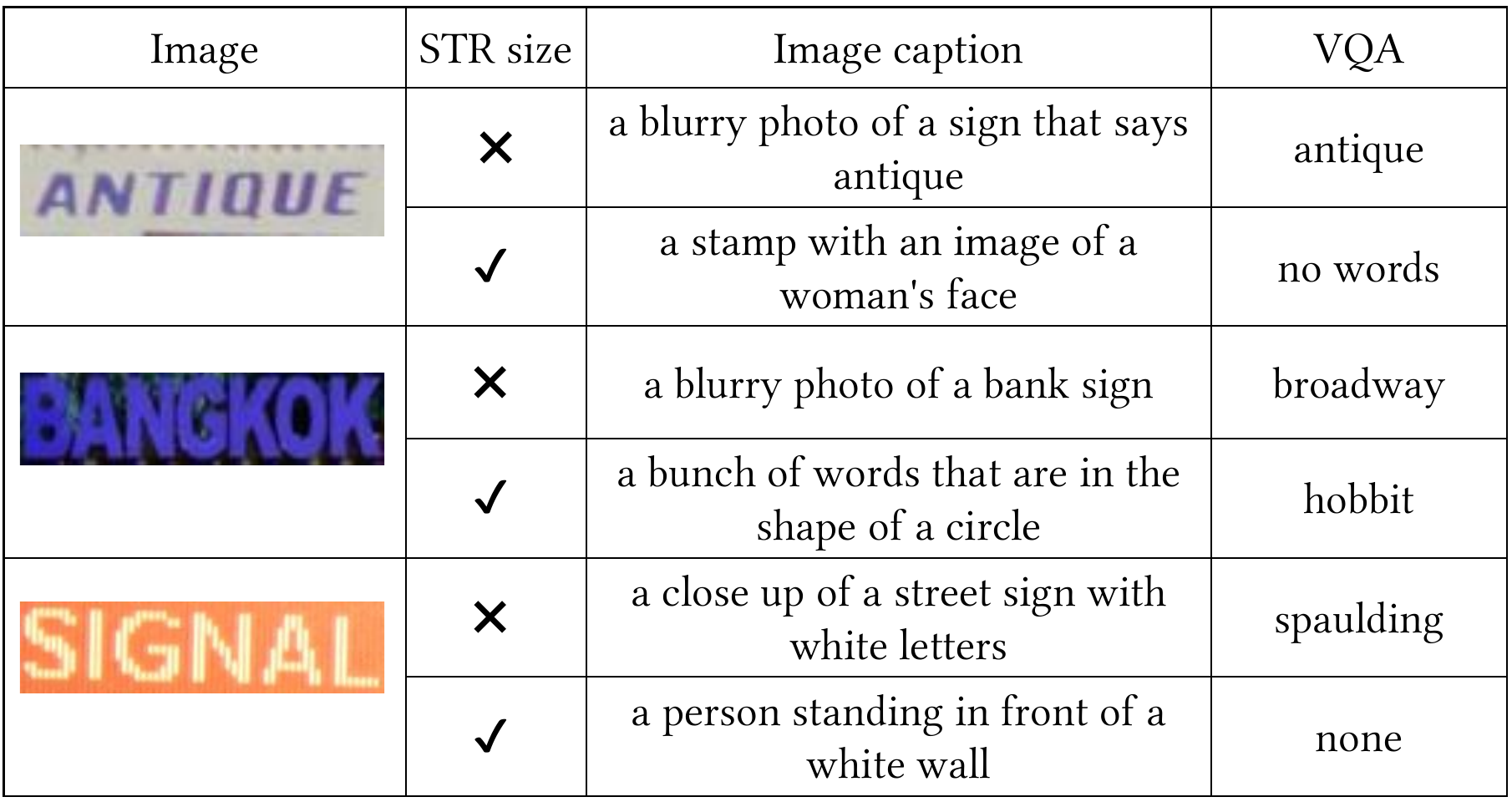}
  \caption{Discussion with Image caption and VQA results on recognition images. Results are predicted by pretrained BLIP model \cite{li2022blip}. STR size means whether resize the input image to 32$\times$ 128. The question for VQA is "Which word is in the image?"}
  \Description{Difference between previous methods and our CLIP-OCR.}
  \label{img_blip}
\end{figure}

In this section, we compare our CLIP-OCR with other image-to-text frameworks, including image caption and VQA.
As illustrated in Fig.\ref{img_blip}, there are two main superiorities for our CLIP-OCR.
First, our SDS establishes an entirely accurate input-output pair for the recognition process by directly sending the ground truth to the text encoder.
But image caption and VQA models often generate inaccurate results which may disturb the learning process ("Bangkok" and "signal" in Fig.\ref{img_blip}).
Second, image caption and text VQA models require high image size but most recognition training frameworks use a small image size with a large batch size.
For example, BLIP \cite{li2022blip} needs 384 $\times$ 384 for image caption and 480 $\times$ 480 for VQA, which results in unacceptable resources cost.
If we manually resize the input to 32 $\times$ 128 for distillation (\checkmark for STR size), their performance will drop severely and can not generate valuable information for guidance while the CLIP model still can align them together (see Fig.\ref{img_clip}).
Therefore, our CLIP-OCR provides the first suitable image-to-text feature flow for STR.
And we believe that regarding CLIP as a progressive guidance flow brings a novel insight for further exploring the potential of the CLIP model for downstream tasks.

\subsection{Limitations}
As we employ feature distillation in the decoding process, for the model with only one decoder layer, we need to the 
replicate the decoder layer by 4 times which may lead to additional computation costs.
However, most methods adopt lightweight decoder which does not require much computation cost.
As shown in Table.\ref{tab:sota}, our CLIP-OCR only has 31.1M parameters which are comparable with MGP-Small and MaskOCR-Small.

\section{Conclusion}
In this paper, we focus on exploring the linguistic knowledge in CLIP and propose a novel Symmetrical Linguistic Feature Distillation (CLIP-OCR) for scene text recognition. 
Rather than only using CLIP for visual feature encoding, CLIP-OCR further focus on leveraging the linguistic knowledge in CLIP by introducing a symmetrical distillation strategy (SDS) and Linguistic Consistency Loss (LCL).
Firstly, SDS combines CLIP image and text encoder symmetrically to generate an image-to-text guidance flow with both visual and linguistic knowledge.
Secondly, we design LCL to improve the learning efficiency of linguistic knowledge by aligning second-order statistics.
Overall, with alignment property in CLIP, we make a first attempt to bridge the transition from image to text progressively for recognition which also provides a new insight 
for exploring CLIP with a unidirectional feature flow.
Extensive experiments on six benchmarks verify the effectiveness of the proposed CLIP-OCR.
And we will further explore the potential of CLIP for recognition in the future. 

\begin{acks}
  This work is supported by the National Key Research and Development Program of China (2022YFB3104700), the National Nature Science Foundation of China (62121002, 62232006, 62102384).
\end{acks}

\bibliographystyle{ACM-Reference-Format}
\balance
\bibliography{sample-base}

\appendix

\section{Additional Experiments}
\subsection{Time efficiency of CLIP-OCR}
Table \ref{tab:time} shows the time efficiency of our method and other compared methods.
The FPS is calculated by averaging the inference time over 3000 images.
All experiments are employed on an NVIDIA 3090 GPU.
As a result, our model achieves 20.9 ms/img which is satisìed for real-time applications. 
This justyìes that our distillation framework can benefit from the large vision-language pretrain model
without introducing much computation cost. 
Compared with other recognition models, our CLIP-OCR obtains comparable inference time with higher performance, we believe that the significant accuracy improvement in our method is worth the sacriìce of such speed.

\subsection{Selection of Distillation Layers}
As CLIP-OCR introduces guidance on all stages of the recognition model. 
To verify whether introducing guidance on all stages of the recognition model is redundant, we further evaluate the results by distilling on fewer stages.
In Table \ref{tab:distill_layer}, only 1,2, or 4 stages in the recognition decoder are selected for distillation. 
Results show that using fewer stages leads to a larger performance drop.
This is reasonable as distillation on more stages can provide more detailed knowledge and reduce the difficulty
of the optimization.
Therefore, distillation on all stages is a straightforward and effective strategy.

\subsection{Analysis on Non-linguistic Data}
To further evaluate our method on the non-linguistic scene, we sample 1k texts in the IC15 test set and build a synth-shuffle test dataset by shuffling the position of characters in each text randomly. And we also build a synth-random test dataset with 1k samples where the texts are generated by random characters. As shown in Table \ref{tab:non-linguistic}, for synth-shuffle, our method still performs better than the baseline which is benefit from the strong visual feature extraction ability in CLIP. But for synth-random, our method performs slightly lower than the baseline. This is because there is not only no linguistic information, but also the frequency of characters is completely different from the general words. Though our method encounters difficulty with random characters, we believe that it is worth introducing linguistic knowledge since texts in most real-life applications contain much linguistic information.

\begin{table}[!h]
  \caption{Time efficiency of CLIP-OCR.}
  \label{tab:time}
  \begin{tabular}{c | c | c}
    \toprule
     Method & Avg & Time(ms/img)\\
    \midrule
    RobustScanner & 88.4 & 40.0 \\
    VisionLAN & 91.2 & 14.3 \\
    ABINet & 92.3 & 38.6 \\
    Parseq$_A$ &93.3 & \textbf{11.5} \\
    CLIP-OCR & \textbf{93.8} & 20.9 \\
  \bottomrule
\end{tabular}
\end{table}

\begin{table}[!h]
  \caption{Selection of Distillation Layers.}
  \label{tab:distill_layer}
  \begin{tabular}{c | c }
    \toprule
    Distillation layer index & Avg \\
    \midrule
    (1) & 93.1 \\
    (1,3) & 93.2 \\
    (1,2,3,4) & \textbf{93.4} \\
  \bottomrule
\end{tabular}
\end{table}

\begin{table}[!h]
  \caption{Results on Non-linguistic dataset.}
  \label{tab:non-linguistic}
  \begin{tabular}{c | c | c}
    \toprule
     Method &  synth-shuffle & synth-random\\
    \midrule
    Baseline & 87.8 & \textbf{71.4} \\
    CLIP-OCR & \textbf{89.0} & 70.9 \\
  \bottomrule
\end{tabular}
\end{table}

\end{document}